\def\year{2021}\relax
\documentclass[letterpaper]{article} 
\usepackage{aaai21}  
\usepackage{times}  
\usepackage{helvet} 
\usepackage{courier}  
\usepackage[hyphens]{url}  
\usepackage{graphicx} 
\usepackage{natbib}  
\usepackage{caption} 
\title{On the Interpretability of Deep Learning Based Models for Knowledge Tracing}
\author {
    Xinyi Ding,\textsuperscript{\rm 1}
    Eric C. Larson \textsuperscript{\rm 2}\\
}
\affiliations {
    \textsuperscript{\rm 1} ZheJiang GongShang University \\
    \textsuperscript{\rm 2} Southern Methodist University \\
    dxywill@gmail.com, eclarson@lyle.smu.edu
}
\begin{document}

\maketitle

\begin{abstract}
Knowledge tracing allows Intelligent Tutoring Systems to infer which topics or skills a student has mastered, thus adjusting curriculum accordingly. Deep Learning based models like Deep Knowledge Tracing (DKT) and Dynamic Key-Value Memory Network (DKVMN) have achieved significant improvements compared with models like Bayesian Knowledge Tracing (BKT) and Performance Factors Analysis (PFA). However, these deep learning based models are not as interpretable as other models because the decision-making process learned by deep neural networks is not wholly understood by the research community. In previous work, we critically examined the DKT model, visualizing and analyzing the behaviors of DKT in high dimensional space. In this work, we extend our original analyses with a much larger dataset and add 
discussions about the memory states of the DKVMN model. We discover that Deep Knowledge Tracing has some critical pitfalls: 1) instead of tracking each skill through time, DKT is more likely to learn an `ability' model; 2) the recurrent nature of DKT reinforces irrelevant information that it uses during the tracking task; 3) an untrained recurrent network can achieve similar results to a trained DKT model, supporting a conclusion that recurrence relations are not properly learned and, instead, improvements are simply a benefit of projection into a high dimensional, sparse vector space.  Based on these observations, we propose improvements and future directions for conducting knowledge tracing research using deep neural network models. 
\end{abstract}

\section{Introduction}

\noindent Knowledge tracing aims to infer which topics or skills a student has mastered based upon the sequence of responses from a question bank. Traditionally student skills were analyzed using parametric models where each parameter has a semantic meaning, such as Bayesian Knowledge Tracing (BKT) \cite{corbett1994knowledge} and  Performance Factors Analysis (PFA) \cite{pavlik2009performance}. BKT, for example, attempts to explicitly model these parameters and use them to infer a binary set of skills as mastered or not mastered. In this model, the `guess' and `slip' parameters in the BKT model reflect the probability that a student could guess the correct answer and make a mistake despite mastery of a skill, respectively.  On the other hand, deep learning based knowledge tracing models like Deep Knowledge Tracing (DKT)~\cite{piech2015deep} and Key-Value Dynamic Memory Network (DKVMN)~\cite{zhang2017dynamic} have improved performance, but their mechanisms are not well understood because none of the parameters are mapped to a semantically meaningful measure. This diminishes our ability to understand how these models perform predictions and what errors these models are prone to make. 
There have been some attempts to explain why DKT works well \cite{khajah2016deep, xiong2016going}, but these studies treat DKT model more like a black box, without studying the state space that underpins the recurrent neural network. 

In this paper, we ``open the box'' of deep neural network based models for knowledge tracing. We aim to provide a better understanding of the DKT model and a more solid footing for using deep neural network models for knowledge tracing. This work extends our previous work \cite{ding2019deep} using a much larger dataset EdNet \cite{choi2019ednet}. 
 We first visualize and analyze the behaviors of the DKT model in a high dimensional space. We track activation changes through time and analyze the impact of each skill in relation to other skills. Then we modify and explore the DKT model, finding that some irrelevant information is reinforced in the recurrent architecture. Finally, we find that an untrained DKT model (with gradient descent applied only to layers outside the recurrent architecture) can be trained to achieve similar performance as a fully trained DKT architecture. Our findings from the EdNet \cite{choi2019ednet} dataset reinforce the conclusions obtained from  ``ASSISTmentsData2009-2010 (b) dataset” \cite{xiong2016going}. We also discuss and visualize the memory states of DKVMN to better understand the hidden skills discovered by this model. Based on our analyses, we propose improvements and future directions for conducting knowledge tracing with deep neural network models.

\section{Related Work}

Bayesian Knowledge Tracing (BKT) \cite{corbett1994knowledge} was proposed by Corbett \textit{et al.} In their original work, each skill has its own model and parameters are updated by observing the responses (correct or incorrect) of applying a skill.   Performance Factors analysis (PFA) \cite{pavlik2009performance} is an alternative method to BKT and is believed to perform better when each response requires multiple skills. Both BKT and PFA are designed in a way that each parameter has its own semantic meaning. For example, the slip parameter of BKT represents the possibility of getting a question wrong even though the student has mastered the skill. These models are easy to interpret, but suffer from scalability issues and often fail to capture the dependencies between each skill because many elements are treated as independent to facilitate optimization.

Piech \textit{et al.} proposed the Deep Knowledge Tracing model (DKT) \cite{piech2015deep}, which exploits recurrent neural networks for knowledge tracing and achieves significantly improved results. They transformed the problem of knowledge tracing by assuming each question can be associated with a ``skill ID'', with a total of $N$ skills in the question bank. The input to the recurrent neural network is a binary vector encoding of skill ID for a presented question and the correctness of the student's response. The output of the recurrent network is a length $N$ vector of probabilities for answering each skill-type question correctly. The DKT model could achieve \textgreater 80\% AUC on the ASSISTmentsData dataset~\cite{feng2006addressing}, compared with the BKT model that achieves 67\% AUC.  Dynamic Key-Value Memory Network for knowledge tracing (DKVMN) \cite{zhang2017dynamic} uses two memories to encode keys (skills) and responses separately. It allows automatic learning of hidden skills.
The success of DKT and DKVMN demonstrates the possibility of using deep neural networks for knowledge tracing.

Despite the effectiveness of DKT model, its mechanism is not well understood by the research community. Khajah \textit{et al.} investigate this problem by extending BKT \cite{khajah2016deep}. They extend BKT by adding forgetting, student ability, and skill discovery components, comparing these extended models with DKT. Some of these extended models could achieve close results compared with DKT. Xiong \textit{et al.} discover that there are duplicates in the original ASSISTment dataset ~\cite{xiong2016going}. They re-evaluate the performance of DKT on different subsets of the original dataset. Both Khajah and Xiong's work are black box oriented---that is, it is unclear how predictions are performed within the DKT model. In our work, we try to bridge this gap and explain some behaviors of the DKT model.

Trying to understand how DKT works is difficult because the mechanisms of RNNs are not totally understood even in the machine learning community. Even though the recurrent architecture is well understood, it is difficult to understand how the model adapts weights for a given prediction task. One common method used is to visualize the neuron activations. Karpathy \textit{et al.}~\cite{karpathy2015visualizing} provide a detailed analysis of the behaviors of recurrent neural network using character level models and find some cells are responsible for long range dependencies like quotes and brackets. They break down the errors and partially explain the improvements of using LSTM. We use and extend their methods, providing a detail analysis of the behaviors of LSTM in the knowledge tracing setting. We also discuss the memory states of DKVMN model.

\begin{figure}[t]
\centering
\includegraphics[height=4in, width=3in]{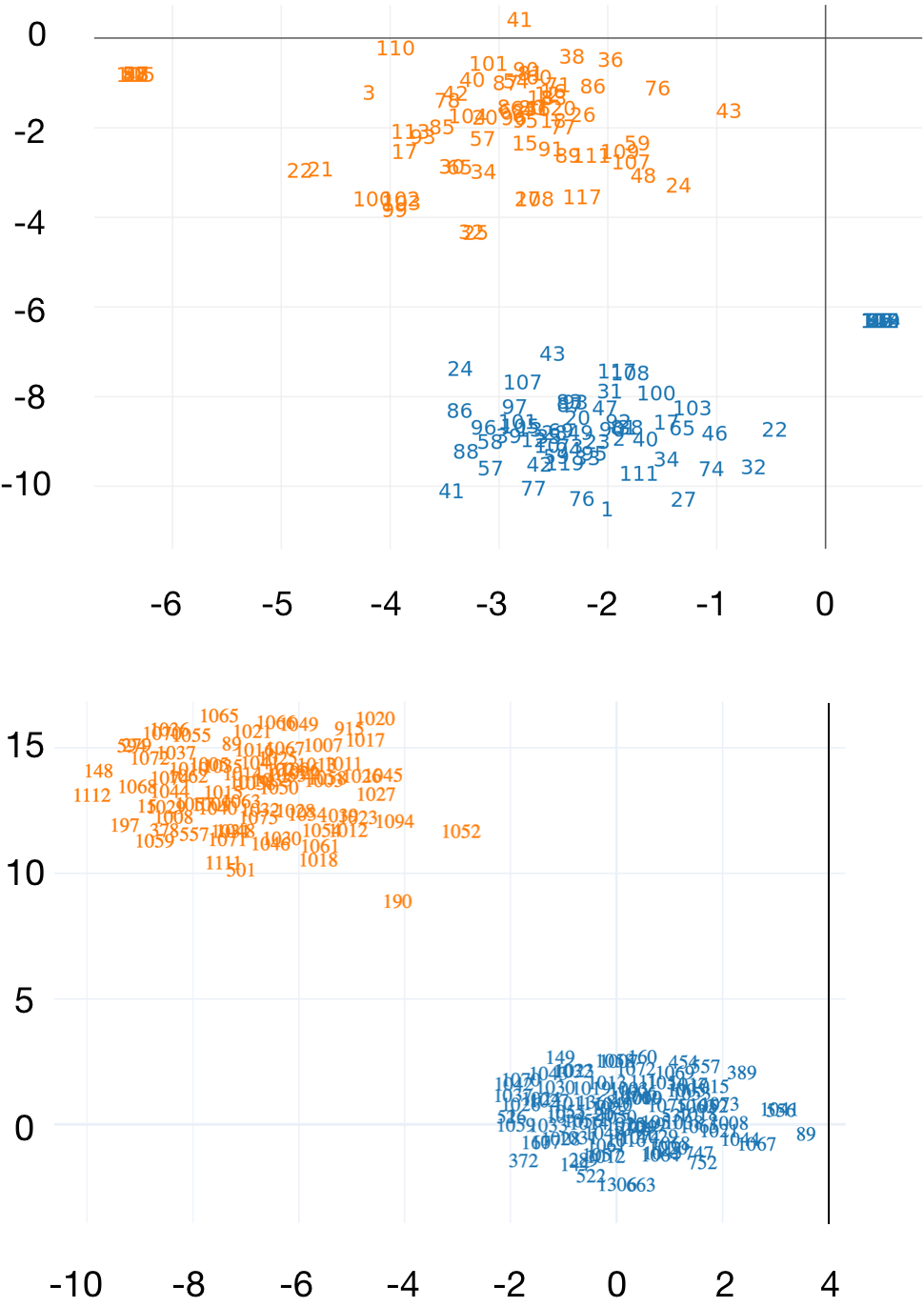}
\caption{ First two components of T-SNE of the activation vector for first time step inputs. Numbers are skill identifiers, blue for correct input, orange for incorrect input. \textbf{TOP:} ASSISTMent dataset \cite{xiong2016going}, \textbf{BOTTOM:} EdNet KT1 \cite{choi2019ednet}}
\label{fig::tsne}
\end{figure}

\section{Deep Knowledge Tracing}

To investigate the DKT model, we perform a number of analyses based upon the activations within the recurrent neural network. We also explore different training protocols and clustering of the activations to help elucidate what is learned by the DKT model. 

\begin{figure*}[t!]
\centering
\includegraphics[height=3.9in, width=6in]{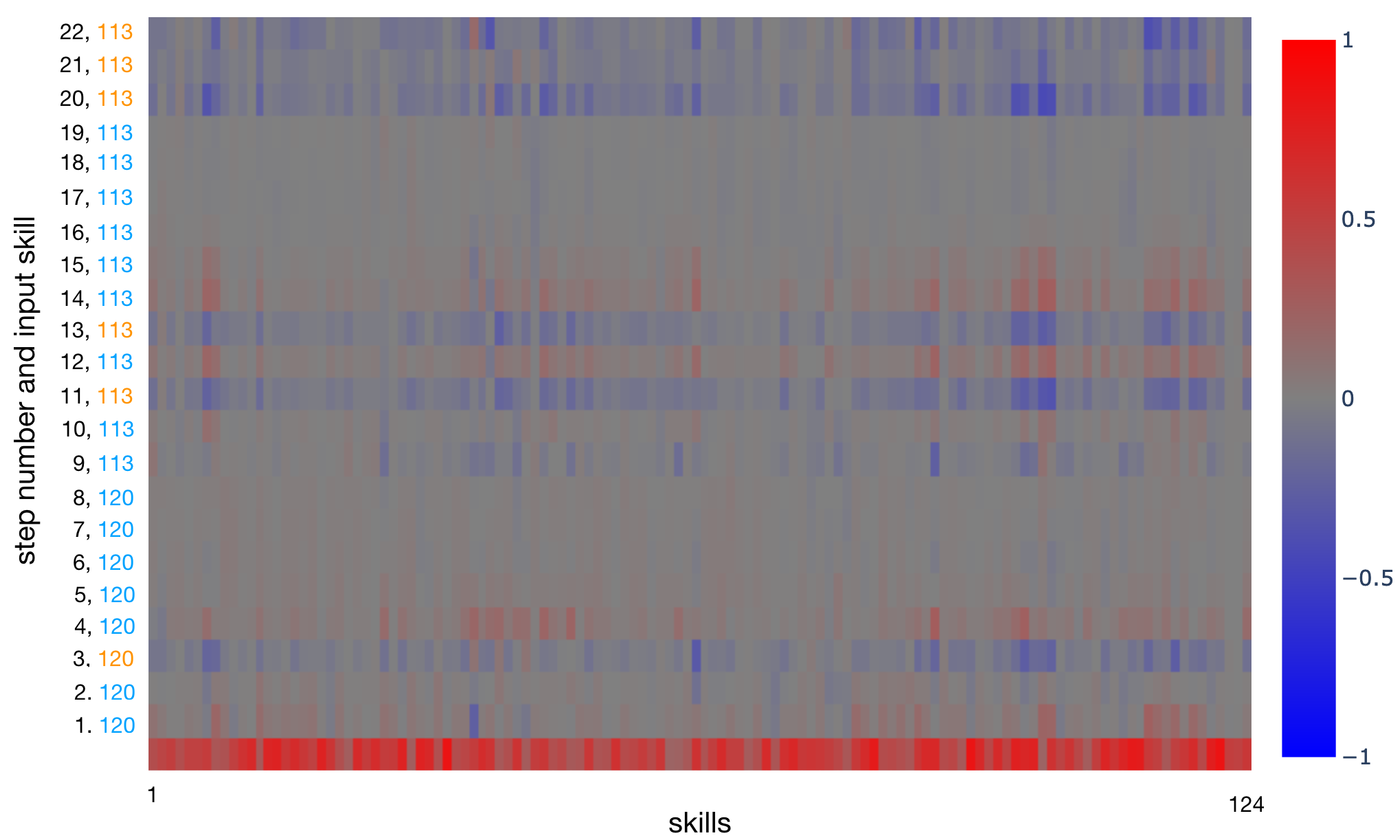}
\caption{The prediction changes for one student, 23 steps, correct input is marked blue, incorrect input is marked orange. ASSISTMent Dataset \cite{xiong2016going}}
\label{fig::skill_impact_1}
\end{figure*}

\begin{figure*}[t!]
\centering
\includegraphics[height=3.9in, width=6in]{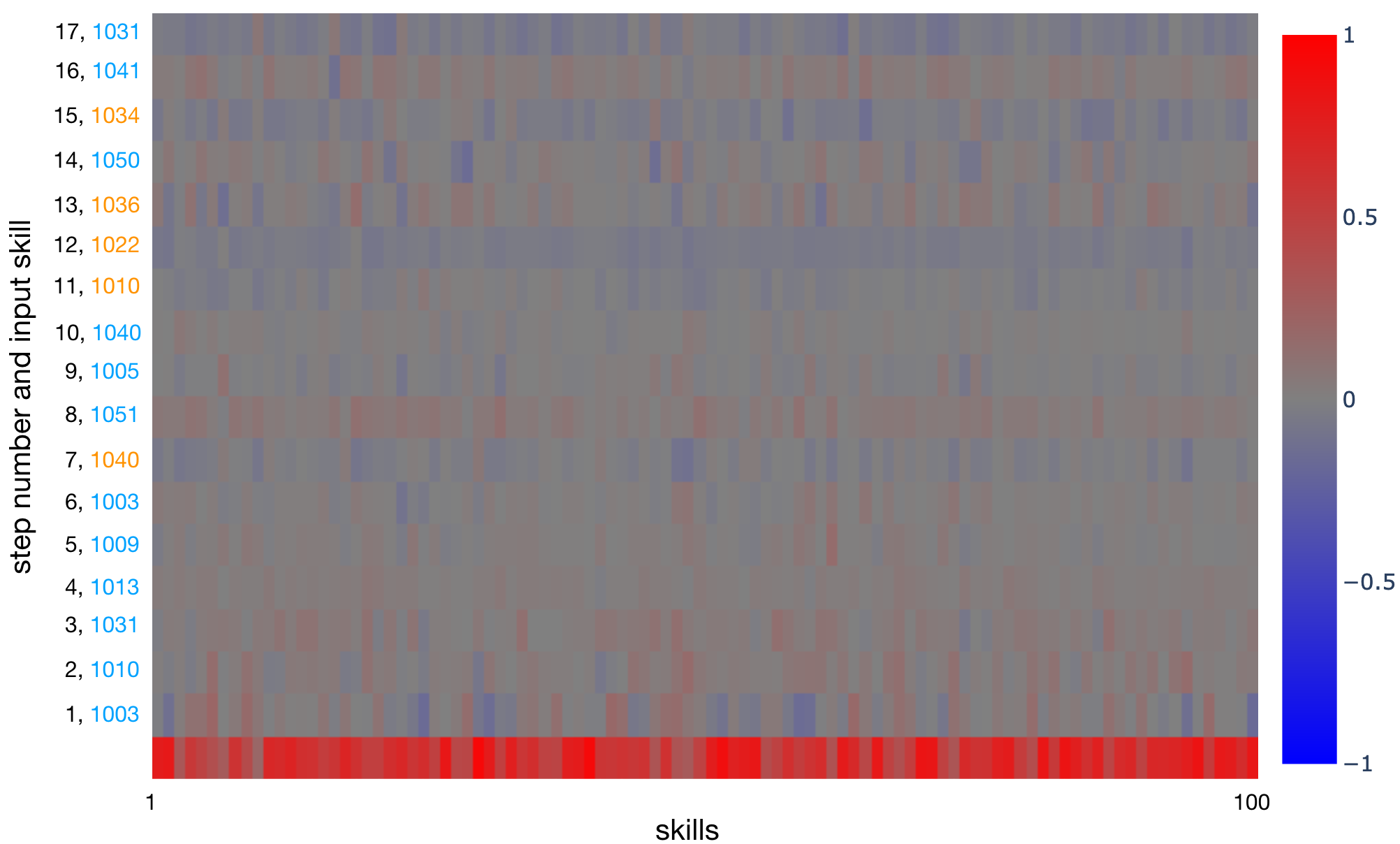}
\caption{The prediction changes for one student, 18 steps, correct input is marked blue, incorrect input is marked orange (only the first 100 skills are showing). EdNet KT1 dataset \cite{choi2019ednet}}
\label{fig::skill_impact_2}
\end{figure*}

\subsection{Experiment setup}

In our original analyses, we used the ``ASSISTmentsData 2009-2010 (b) dataset'' which is created by Xiong \textit{et al.} after removing duplicates \cite{xiong2016going}. In this work, we extend our previous anlyses \cite{ding2019deep} using a much larger dataset EdNet \cite{choi2019ednet} with millions of interactions. The KT1 dataset from EdNet has 188 tags. If one question has multiple tags, these multiple tags will be combined to form a new skill, resulting in 1495 skills in our study.  Like Xiong \textit{et al.}, we also use LSTM units for analysis in this paper. Because we will be visualizing specific activations of the LSTM, it is useful to review the mathematical elements that comprise each unit. An LSTM unit consists of the following parts, where a sequence of inputs $\{ x_1, x_2, ..., x_T \}\in\mathcal{X}$ are ideally mapped to a labeled output sequence $\{ y_1, y_2, ..., y_T \}\in\mathcal{Y}$. The prediction goal is to learn weights and biases ($W$ and $b$) such that the model output sequence ($\{ h_1, h_2, ..., h_T \}\in\mathcal{H}$) is as close as possible to $\mathcal{Y}$:
\begin{equation}
f_t = \sigma (W_f \cdot [h_{t-1}, x_t] + b_f)
\end{equation}
\begin{equation}
i_t = \sigma (W_i \cdot [h_{t-1}, x_t] + b_i)
\end{equation}
\begin{equation}
\tilde{C}_t = \tanh(W_C \cdot [h_{t-1}, x_t] + b_C)
\end{equation}
\begin{equation}
C_t = f_t * C_{t-1} + i_t * \tilde{C}_t
\end{equation}
\begin{equation}
o_t = \sigma (W_o \cdot [h_{t-1}, x_t] + b_o)
\end{equation}
\begin{equation}
h_t = o_t * \tanh(C_t)
\end{equation}

Here, $\sigma$ refers to a logistic (sigmoid) function, $\cdot$ refers to dot products, $*$ refers to element-wise vector multiplication, and $[,]$ refers to vector concatenation. For visualization purposes, we log the above 6 intermediate outputs for each input during testing and concatenate these outputs into a single ``activation'' vector, $a_t = [f_t,i_t,\tilde{C}_t,C_t,o_t,h_t]$. In the DKT model, the output of RNN, $h_t$ is connected to an output layer $y_t$, which is a vector with the same number of elements as skills. We can interpret each element in $y_t$ as an estimate that the student would answer a question from each skill correctly, with larger positive number denoting that the student is more likely to answer correctly and more negative numbers denoting that the student is unlikely to respond correctly. Thus, a student who had mastered all skills would ideally obtain an $y_t$ of all ones. A student who had mastered none of the skills would ideally obtain an $y_t$ of all negative ones.

Deep neural networks usually work in high dimensional space and are difficult to visualize.  Even so,  dimensionality reduction techniques can help to identify clusters.  
For example, Figure \ref{fig::tsne} plots the first two reduced components (using t-SNE~\cite{maaten2008visualizing}) of the activation vector, $a_t$, at the first time step ($t=0$) for a number of different students. The numbers in the plot are skill identifiers. We use color blue to denote a correct response and the color orange to denote an incorrect response. From reducing the dimensionality of the $a_t$ vector for each student, we can see that the activations show a distinct clustering between whether the questions were answered correctly or incorrectly. We might expect to observe sub-clusters of the skill identifiers within each of the two clusters but we do not. This observation supports the hypothesis that correct and incorrect responses are more important for the DKT model than skill identifiers. However, perhaps this lack of sub-clusters is inevitable because we are only visualizing the activations after one time step---this motivates the analysis in the next section.

\subsection{Skill relations}

\begin{figure}[t]
\centering
\includegraphics[height=2.5in, width=3in]{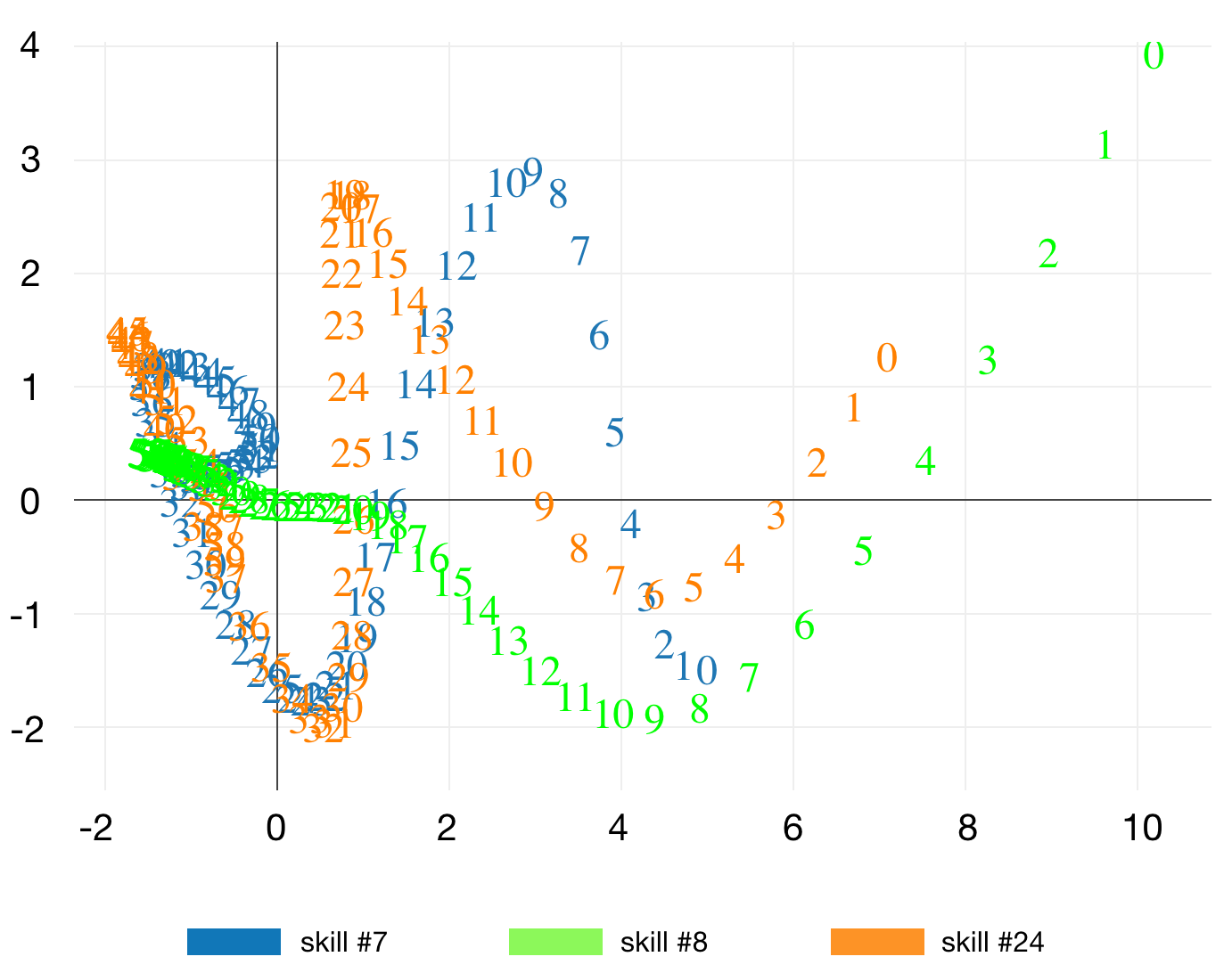}
\caption{Activation vector changes for 100 continuous correctness of randomly picked 3 skills }
\label{fig::oracle_1}
\end{figure}

\begin{figure}[t]
\includegraphics[height=2.5in, width=3in]{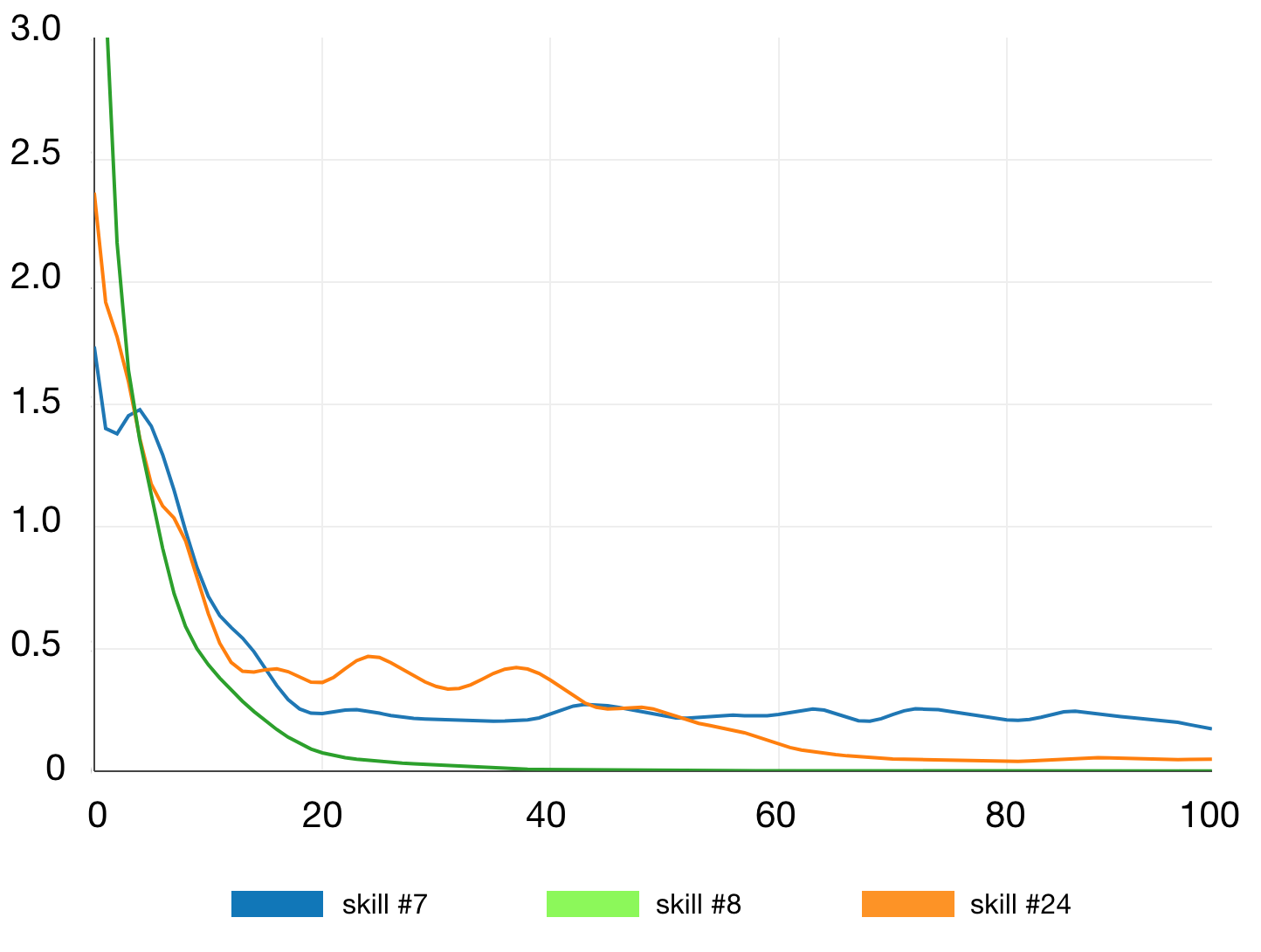}
\caption{Activation vector difference of randomly picked 3 skills through time}
\label{fig::oracle_diff}
\end{figure}

\begin{figure*}
\centering
\includegraphics[height=2.15in, width=6in]{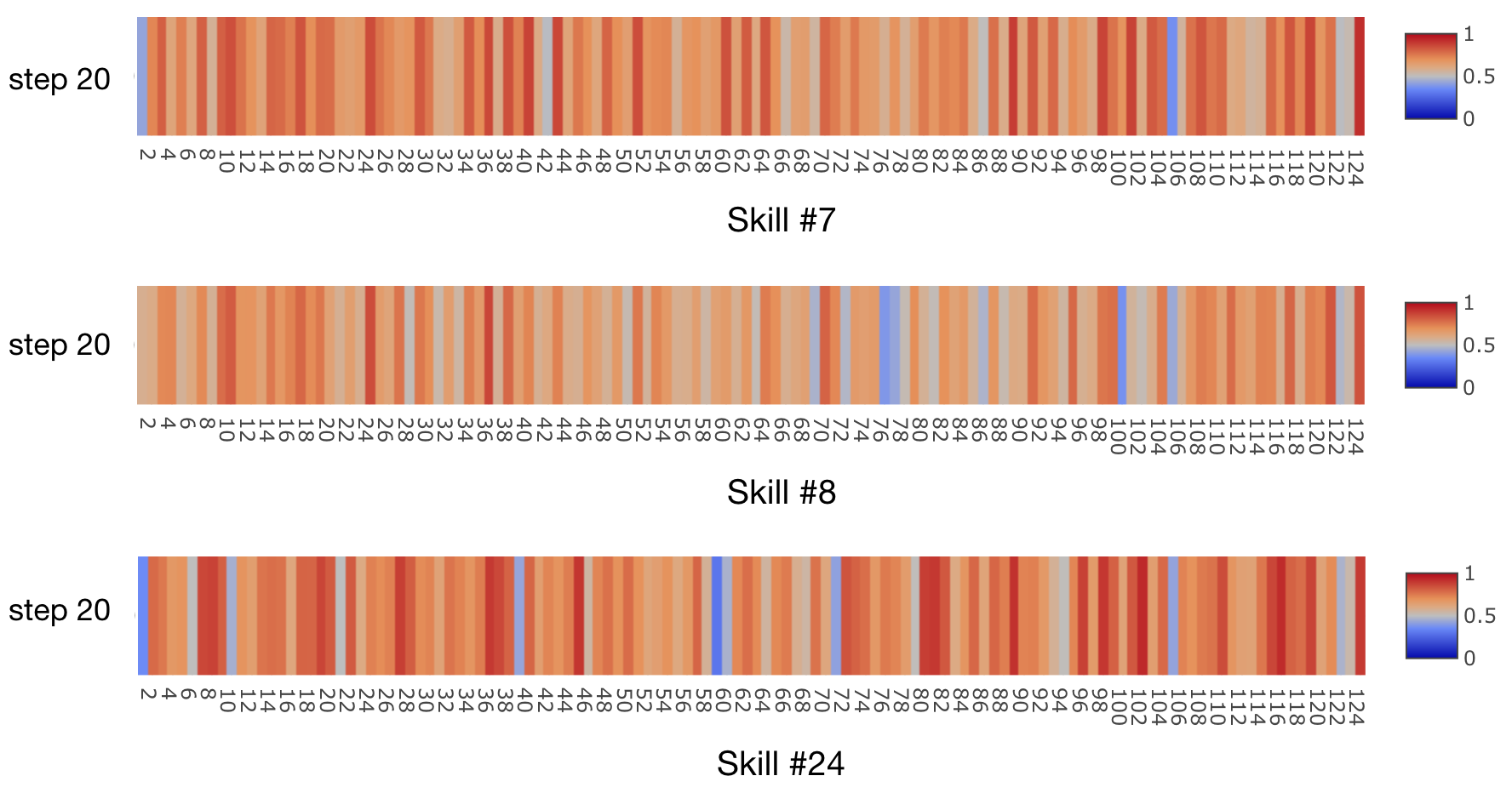}
\caption{Prediction vector after 20 steps for skill \#7, \#8, \#24}
\label{fig::steps_combined}
\end{figure*}

In this section, we try to understand how the prediction vector of one student changes as a student answers more questions from the question bank.  Figure \ref{fig::skill_impact_1}  and Figure \ref{fig::skill_impact_2} plot the prediction difference (current prediction vector - previous prediction vector) for each question response from one particular student (steps are displayed vertically and can be read sequentially from bottom to top). The horizontal axis denotes the skill identifier and the color of the boxes in the heatmap denote the change in the output vector $y_t$.  The initial row in the heatmap (bottom) is the starting values for $y_t$ for the first input. As we can see, if the student answers correctly, most of the $y_t$ values increase (warm color). When an incorrect response occurs, most of the predictions decreases (cold color). This makes intuitive sense. We expect a number of skills to be related so correct responses should add value and incorrect responses should subtract value. We can further observe that changes in the $y_t$ vector diminish if the student correctly or incorrectly answers a question from the same skill several times repeatedly. For example, in figure \ref{fig::skill_impact_1}, observe from step 14 to step 19, where the student correctly answers questions from skill \#113---eventually the changes in $y_t$ come to a steady state.  However, occasionally, we can also notice, a correct response will result in decreases in the prediction vector (observe step 9). This behavior is difficult to justify from our experience, as correctly answering a question should not decrease the mastery level of other skills. Yeung \textit{et al.} have similar findings when investigating single skills \cite{yeung2018addressing}. Observe also that step 9 coincides with a transition in skills being answered (from skill \#120 to \#113). Even so, it is curious that switching from one skill to another would decrease values in $y_t$ even when the response is correct. We also notice this kind of behaviors are consistent across datasets with different size. From this observation, one potential way to improve the DKT model could be adding punishment for such unexpected behaviors (for example, in the loss function of the recurrent network).

\subsection{Simulated data}

From the above analysis, we see in figure \ref{fig::skill_impact_1}, from step 14 to step 19, the student correctly answers question from skill \#113 and the changes in $y_t$ diminish---perhaps an indication that the vector is converging. Also, we see that for each correct input, most of the elements of $y_t$ increase by some margin, regardless of the input skill.  To have a better understanding of this convergence behavior, we simulate how the DKT model would respond to an \textit{Oracle Student}, which will always answer each skill correctly. We simulate how the model responds to the \textit{Oracle Student} correctly answering 100 questions from one skill. We repeat this for three randomly selected skills.  

We plot the convergence of each skill using the activation vector $a_t$ reduced to a two-dimensional plot using t-SNE (Figure \ref{fig::oracle_1}). The randomly chosen skills were \#7. \#8, and \#24. 
As we can see, each of the three skills starts from a different location in the 2-D space. However, they each converges to near the same location in space. In other words, it seems DKT is learning one ``oracle state'' and this state can be reached by practicing any skill repeatedly, regardless of the skill chosen. We verified this observation with a number of other skills (not shown) and find this behavior is consistent. Therefore, we hypothesize that DKT is learning a `student ability' model, rather than a `per skill' model like BKT. To make this observation more concrete, in Figure \ref{fig::oracle_diff} we plot the euclidean distance between the current time step activation vector, $a_t$, and the previous activations, $a_{t-1}$, we can see the difference becomes increasingly small after 20 steps. Moreover, the euclidean distance between each activation vector learned from each skill becomes extremely small, supporting the observation that not only is the $y_t$ output vector converging, but all the activations inside the LSTM network are converging. We find this behavior curious because it means that the DKT model is not remembering what skill was used to converge the network to an `oracle state.' Remembering the starting skill would be crucial for predicting future performance of the student, yet the DKT model would treat every skill identically. We also analyzed a process where a student always answers responses incorrectly and found there is a similar phenomenon with convergence in an anti-oracle state.


Figure \ref{fig::steps_combined} shows the skills prediction vector after answering correctly 20 times in a row. We can see the predictions of most skills are above 0.5, regardless of the specific practice skill used by the \textit{Oracle Student}. Thus, we believe that the DKT model is not really tracking the mastery level of \textit{each skill}, it is more likely learning an `ability model' from the responses. Once a student is in this oracle state, DKT will assume that he/she will answer most of the questions correctly from any skill. We hypothesize that this behavior could be mitigated by using an ``attention'' vector during the decoding of the LSTM network \cite{vaswani2017attention}. Self attention in recurrent networks decodes the state vectors by taking a weighted sum of the state vectors over a range in the sequence (weights are dynamic based on the state vectors). For DKT, this attention vector could also be dynamically allocated based upon the skills answered in the sequence, which might help facilitate remembering long-term skill dependencies.     

\subsection{Temporal impact}
RNNs are typically well suited for tracking relations of inputs in a sequence, especially when the inputs occur near one another in the sequence. However, long range dependencies are more difficult for the network to track \cite{vaswani2017attention}. In other words, the predictions of RNN models will be more impacted by recent inputs. For knowledge tracing, this is not a desired characteristic.  Consider two scenarios as shown below: For each scenario, the first line is the skill numbers and the second line are responses (1 for correctness and 0 for incorrectness). Both two scenarios have the same number of attempts for each skill (4 attempts for skill \#9, 3 attempts for skill \#6 and 2 attempts for skill \#24). Also, the ordering of correctness within each skill is the same (\textit{e.g.}, 1, 0, 0, 0 for skill \#9). 

\begin{table}[htbp]
\centering
\label{tab::origins}
\begin{tabular}{|l|l|l|l|l|l|l|l|l|l|}
\hline
\multicolumn{10}{|c|}{Scenario \#1}\\
\hline
Skill ID &6 &6 &9 &9 &9 &9 &\textbf{24} &24 &6\\
Correct &1 &1 &1 &0 &0 &0 &0 &0 &1\\ 
\hline \hline
\multicolumn{10}{|c|}{Scenario \#2}\\
\hline
Skill ID &9 &9 &9 &9 &6 &6 &6 &\textbf{24} &24\\
Correct &1 &0 &0 &0 &1 &1 &1 &0 &0\\ 
\hline
\end{tabular}
\end{table}

For models like BKT, there is a separate model for each skill. Thus, the relative order of different skills presented has no influence, as long as the ordering within each skill remains the same. In other words, for each skill the ordering of correct and incorrect attempts remains the same, but different skills can be shuffled into the sequence. For BKT, it will learn the same model from these two scenarios, but it may not be the case for DKT. The DKT model is more likely to predict incorrect response after seeing three incorrect inputs in a row because it is more sensitive to recent inputs in the sequence. This means, for the first scenario, first attempt of skill \#24 (in bold) will be more likely predicted incorrect because it follows three incorrect responses. For the second scenario, first attempt of skill \#24 (in bold) is more likely to be predicted correct. Thus the DKT model might perform differently on the given scenarios. 


Khajah \textit{et al.} also alluded to this recency effect in \cite{khajah2016deep}. In this paper, we examine this phenomenon in a more quantitative way. We shuffle the dataset in a way that keeps the ordering within each skill the same, but spreads out the responses in the sequence. This change should not change the prediction ability of models like BKT. The results are shown in Table \ref{tab::auc} and Table \ref{tab::r2} using standard evaluation criteria for this dataset. All results are based on a five-fold cross validation of the dataset. When comparing DKT on the original dataset to the ``spread out'' dataset ordering, we see that the relative ordering of skills has significant negative impact on the performance of the model. From these observations, we see the behaviors of DKT is more like PFA which counts prior frequencies of correct and incorrect attempts other than BKT and the design of the exercises could have a huge impact on the model (For example, the arrangements of easy and hard exercises).

\subsection{Is the RNN representation meaningful? }

Recurrent models have been successfully used in practical tasks like natural language processing \cite{devlin2018bert}. These models can take days or even weeks to train. Wieting \textit{et al.} \cite{wieting2019no} argue that RNNs might not be learning a meaningful state vector from the data. They show that a randomly initialized RNN model (with only $W_o$ and $b_o$ trained) can achieve similar results to models where all parameters are trained. This result is worrying because it may indicate that the RNN performance is due mostly to simply mapping input data to random high dimensional space. Once projected into the random vector space linear classification can perform well because points are more likely to be separated in a sparse vector space. The actual vector space may not be meaningful. We perform a similar experiment in training the DKT model. We randomly initialize the DKT model and only train the last linear layer ($W_o$ and $b_o$) that maps the output of LSTM  $h_t$ to the skill vector, $y_t$. As shown in Table \ref{tab::auc} and Table \ref{tab::r2}, the untrained recurrent network performs similarly to the trained network.   

\begin{table}[h]
\centering
\caption{Area under the ROC curve}
\label{tab::auc}
\begin{tabular}{|l|l|l|l|l|l|}
\hline
           & PFA  & BKT  & DKT  & \begin{tabular}[c]{@{}l@{}}DKT\\ (spread)\end{tabular} & \begin{tabular}[c]{@{}l@{}}DKT\\ (untrained)\end{tabular} \\ \hline
09-10 (a)  & 0.70 & 0.60 & 0.81 & 0.72                                                    & 0.79                                                   \\ \hline
09-10 (b)  & 0.73 & 0.63 & 0.82 & 0.72                                                    & 0.79                                                   \\ \hline
09-10 (c)  & 0.73 & 0.63 & 0.75 & 0.71                                                    & 0.73                                                   \\ \hline
14-15      & 0.69 & 0.64 & 0.70 & 0.67                                                    & 0.68                                                   \\ \hline
KDD        & 0.71 & 0.62 & 0.79 & 0.76                                                    & 0.76                                                   \\ \hline
EdNet     &      &      & 0.70 & 0.68         & 0.67            \\ \hline
\end{tabular}
\end{table}

\begin{table}[h]
\centering
\caption{Square of linear correlation ($r^2$) results}
\label{tab::r2}
\begin{tabular}{|l|l|l|l|l|l|}
\hline
           & PFA  & BKT  & DKT  & \begin{tabular}[c]{@{}l@{}}DKT\\ (spread)\end{tabular} & \begin{tabular}[c]{@{}l@{}}DKT\\ (untrained)\end{tabular} \\ \hline
09-10 (a)  & 0.11 & 0.04 & 0.29 & 0.15                                                    & 0.25                                                   \\ \hline
09-10 (b)  & 0.14 & 0.07 & 0.31 & 0.14                                                    & 0.26                                                   \\ \hline
09-10 (c)  & 0.14 & 0.07 & 0.18 & 0.14                                                    & 0.15                                                   \\ \hline
14-15      & 0.09 & 0.06 & 0.10 & 0.08                                                    & 0.09                                                   \\ \hline
KDD        & 0.10 & 0.05 & 0.21 & 0.17                                                    & 0.17                                                   \\ \hline
EdNet     &      &      & 0.11 & 0.09       & 0.08            \\ \hline
\end{tabular}
\end{table}

\section{Dynamic Key-Value Memory Network}

\begin{figure*}
\centering
\includegraphics[height=2.15in, width=6in]{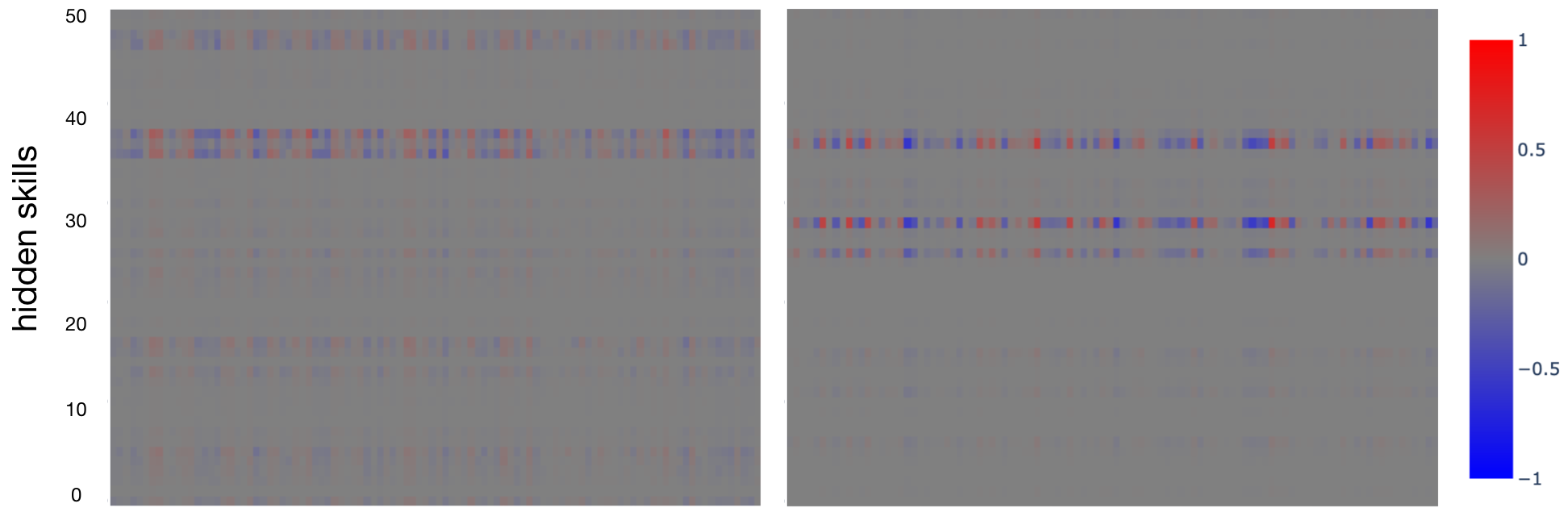}
\caption{Memory changes (current memory value - previous memory value) for different question inputs. \textbf{LEFT:} compact skill ID 1009 (tags: 77), \textbf{RIGHT:} compact skill ID 1078 (tags: 77;179)}
\label{fig::mem_state}
\end{figure*}

\noindent Dynamic Key-Value Memory Network for knowledge tracing (DKVMN) \cite{zhang2017dynamic}  has one static key memory $M^k$, which stores the encodings of all skills. The content of $M^k$ does not change with time.  DKVMN also has one dynamic value memory $M_t^v$ for storing the current states of corresponding skills. The content of $M_t^v$ is updated after each response. There are two stages involved in the DKVMN model. In the read stage, a query skill $q_t$ is first embedded to get $k_t$. Then a correlation weight is calculated:
\begin{equation}
    w_t(i) = \textit{softmax}(k_t^TM^k(i))
\end{equation}
The current state of skill $q$ is thus calculated as follows:
\begin{equation}
    r_t = \sum w_t(i)M_t^v(i)
\end{equation}
The authors concatenate the query skill $q_t$ with $r_t$ to get the final output $p_t$ arguing that the difficult level of each skill might be different. The second stage is to update the memory network $M_t^v$. The embedding of the combination of the skill query $q_t$ and the actual correctness $r_t$ is used to create an erase vector $e_t$ and an add vector $a_t$. The new value matrix is updated using the following equations:
\begin{equation}
    \tilde{M}_t^v(i) = M_{t-1}^v(i)[\textbf{1} - w_t(i)e_t]
\end{equation}
\begin{equation}
    M_t^v(i) = \tilde{M}_t^v(i) + w_t(i)a_t
\end{equation}

The assumption behind the DKVMN model is for each question, there are some hidden knowledge components (skills) governing the response. All hidden skills are encoded as the $M^k$. Users decide beforehand how many skills are there for a given dataset and the model will learn to discover the hidden skills. Figure \ref{fig::mem_state} shows the memory changes (current memory value - previous memory value) for different inputs. The memory size shown in the figure is limited to 50 for display purpose. For one specific question, we observe no matter it's a correct response or incorrect response, the same locations in the memory are activated. This meets our intuition that some hidden skills are responsible for one specific question. We also observe that the size of the memory does not have too much impact on the performance of the model. If we change to use a much larger memory size, which means to use more hidden skills, we can see different positions are activated. But for the questions with the same skill id, the activated positions are the same. Thus, even though DKVMN model can learn hidden skills automatically, we still do not know the ideal number of hidden skills. As long as the memory size is not too small, this model can always learn a reasonable hidden skill set. These discovered skills may or may not map to the skills discovered by human experts.

Figure \ref{fig::mem_state} left gives the value memory changes for the compact skill 1009, which consists of tag 77. Figure \ref{fig::mem_state} right gives the value memory changes for skill 1078 which consists of tag 77 and 179. We assumed the activation hidden skills of skill 1078 might contains all locations of skill 1009 (since it requires tag 77 and tag 179). However, despite experimenting with different memory sizes, we did not observe this relationship.

\section{Conclusion and Future Work}
This work extended our previous work \cite{ding2019deep} using a much larger dataset EdNet \cite{choi2019ednet}. Using this new data, we dive deep into the Deep Knowledge Tracing model, finding similar conclusions. Using dimensionality reduction and temporal sequence behavior, we find that the DKT model is most likely learning an `ability' model, rather than tracking each individual skill. Moreover DKT is significantly impacted by the relative ordering of skills presented. We also discover that a randomly initialized DKT with only the final linear layer trained achieves similar results to the fully trained DKT model. In other words, the DKT model performance gains may stem from mapping input sequences into a random high dimensional vector space where linear classification is easier because the space is sparse.  We also discussed the memory states of DKVMN model. Several mitigating measures are suggested in this paper, including the use of a loss function that mitigates unwanted behaviors and the use of an attention model to better capture long term skill dependencies.   

\bibliography{main}
\end{document}